\documentclass[pdflatex,sn-mathphys-num]{sn-jnl}

\usepackage{graphicx}%
\usepackage{multirow}%
\usepackage{amsmath,amssymb,amsfonts}%
\usepackage{amsthm}%
\usepackage{mathrsfs}%
\usepackage{xcolor}%
\usepackage{textcomp}%
\usepackage{manyfoot}%
\usepackage{booktabs}%
\usepackage{algorithm}%
\usepackage{algorithmicx}%
\usepackage{algpseudocode}%
\usepackage{listings}%

\usepackage{subcaption}

\usepackage[notheorems]{changdefs}
\usepackage{color}

\theoremstyle{thmstyleone}%
\newtheorem{theorem}{Theorem}
\newtheorem{proposition}[theorem]{Proposition}%

\theoremstyle{thmstyletwo}%
\newtheorem{remark}{Remark}%

\theoremstyle{thmstylethree}%

\begin{document}

\title{Generative structure search for efficient and diverse discovery of molecular and crystal structures}


\author[1]{\fnm{Yifang} \sur{Qin}}\email{qinyifang@pku.edu.cn}

\author[2]{\fnm{Yu} \sur{Shi}}\email{shiyu@bza.edu.cn}

\author[3]{\fnm{Junfu} \sur{Tan}}\email{jtan370@gatech.edu}


\author*[2]{\fnm{Chang} \sur{Liu}}\email{liuchang@bza.edu.cn}

\author*[1]{\fnm{Ming} \sur{Zhang}}\email{mzhang\_cs@pku.edu.cn}

\author*[2,4]{\fnm{Ziheng} \sur{Lu}}\email{zluag@connect.ust.hk}

\affil[1]{\orgdiv{School of Computer Science}, \orgname{Peking University}
\orgaddress{\city{Beijing}, \postcode{100871}, \country{China}}}

\affil[2]{\orgname{Zhongguancun Academy}, \orgaddress{\city{Beijing}, \postcode{100094}, \country{China}}}

\affil[3]{\orgdiv{School of Electrical and Computer Engineering}, \orgname{Georgia Institute of Technology}, \orgaddress{\city{Atlanta}, \postcode{30332}, \state{Georgia}, \country{USA}}}

\affil[4]{\orgname{Kairos Materials}, \orgaddress{\city{Beijing}, \postcode{100094}, \country{China}}}


\abstract{
Predicting stable and metastable structures is central to molecular and materials discovery, but remains limited by the cost of searching high-dimensional energy landscapes. Deep generative models offer efficient structure sampling, yet their outputs remain shaped by training data and can underexplore minima that are rare but physically relevant. We introduce generative structure search (GSS), a unified framework that formulates diffusion-based generation and random structure search (RSS) as limiting regimes of a common sampling process driven by learned score fields and physical forces. Coupling these drivers lets GSS use data priors to accelerate sampling while retaining energy-guided exploration of local minima. Across molecular and crystalline systems, GSS recovers diverse metastable structures with more than tenfold lower sampling cost than RSS for broad coverage and remains effective for compositions outside the training distribution. The results establish a physically grounded generative search strategy for discovering structures beyond the reach of data-driven sampling alone.
}

\maketitle

\section{Introduction}\label{sec1}

Searching for stable structures of molecules and materials is the first step toward predicting their physical and chemical properties through computational methods.
This task is central to computational drug design, where molecular conformations govern binding affinity and bioavailability, as well as to materials discovery, where polymorphic phases determine mechanical, electronic, and catalytic properties.
It is well-known that the structure for a given material composition or a molecular graph is not unique, and different structures lead to different properties.
Formally, stable structures correspond to local minima on the potential energy surface (PES) that are energetically competitive under relevant thermodynamic conditions.

Due to the immense structure space and complicated PES, finding stable structures could be intricate and costly.
A variety of computational methods have been developed to tackle this challenge, including simulated annealing~\cite{pannetier1990prediction}, basin hopping~\cite{wales1997global}, evolutionary algorithms~\cite{oganov2006crystal}, and metadynamics~\cite{martovnak2003predicting}.
Among them, random structure search (RSS)~\cite{pickard2006high,pickard2011ab} stands out for its simplicity and unbiased nature: it starts the exploration from randomly generated seed structures, then undergoes a relaxation process by minimizing the potential energy.
Despite the variety of strategies, all these methods share a common bottleneck---they require a large number of expensive energy evaluations to sufficiently explore the configuration space. In particular, due to the locality of the optimization process, RSS may require a large amount of trials to meet a seed structure that falls into the basin of a stable structure.

\begin{figure*}[t!]
    \centering
    \includegraphics[width=0.88\textwidth]{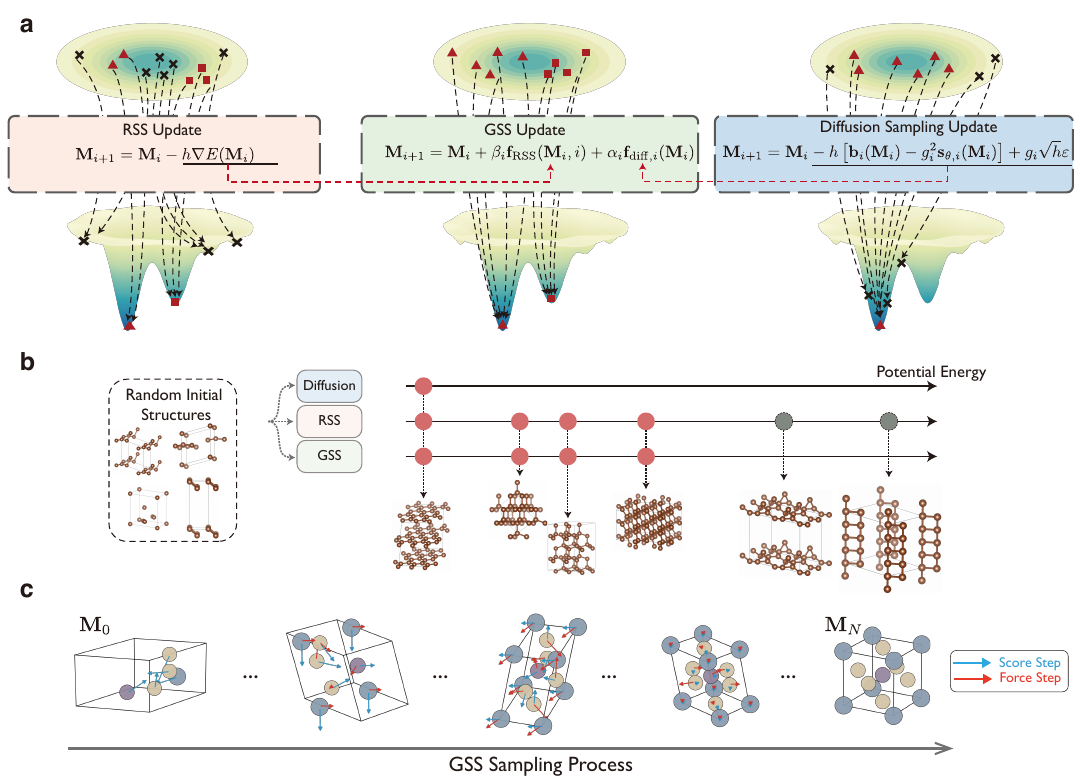}
    \caption{\textbf{a}, Overview of the proposed generative structure search (GSS) approach (middle) in comparison with random structure search (RSS; left) and diffusion generative model (right).
    All three methods generate structures from randomly sampled initial configurations. RSS undergoes a relaxation process by minimizing energy, converging to local minima, but a significant portion of which (the black crosses) may have high energy. Diffusion model consecutively transforms the structure using a learned diffusion process targeting the global minimum (the red triangles), but lacks coverage on competitive stable structures (the red squares), and may end up with arbitrary unstable structures due to generalization error and lack of physical information. In contrast, GSS properly combines the generation processes from the two sides, where the relaxation force biases intermediate structures towards other local minima and ensures they end up there (covering the red squares), and the diffusion update provides a global guide towards low-energy regions (low portion of black crosses). 
    \textbf{b}, Illustration of the GSS sampling process. Starting from a randomly initialized structure $\mathbf{M}_0$, the atomic score (blue arrows) and relaxation force (red arrows) jointly guide the structure through intermediate noisy states toward a clean, stable configuration $\mathbf{M}_N$.
    \textbf{c}, Comparison of the three methods on elemental carbon. Random initial structures (left) are sampled by each method independently. On the energy axis (right), diffusion concentrates on graphite alone, whereas both RSS and GSS recover all four reference polymorphs (graphite, diamond, lonsdaleite, bct-C4). RSS, however, additionally produces invalid high-energy structures (gray dots), which are largely eliminated by GSS.
    }
    \label{fig:framework}
\end{figure*}

On the other side, recent remarkable advances of artificial intelligence (AI) methods have shown their power in solving science problems. Particularly relevant to structure search, diffusion-based generative models~\cite{sohl2015deep,ho2020denoising,song2021score} have shown remarkable capability in generating high-dimensional continuous variables, encouraging its application in generating molecular structures~\cite{xu2022geodiff,zheng2024predicting,abramson2024accurate,yang2024mattersim,zeni2025generative}, which shows an unprecedented efficiency in finding desired structures on new systems.
Nevertheless, these models are trained on a fixed distribution of demonstrated structures that concentrates on limited portion of low-energy region of the configuration space, and in many cases targets only the single global-energy-minimum structure per system. This inherent distributional bias limits their ability to explore alternative metastable structures that are energetically competitive yet geometrically distinct from the training modes.
For instance, as shown in \figref{framework}c, a diffusion model applied to elemental carbon recovers only graphite while missing diamond, lonsdaleite, and bct-C4.
Moreover, for systems not represented in the training data, diffusion models rely solely on statistical generalization, and the generated structures may resemble known geometries but lack a physical guarantee of stability.

In this work, we set out the endeavor to combine the best of the two worlds by proposing the Generative Structure Search (GSS) approach, 
which leverages the efficient generation process of a diffusion model as an oracle guide 
towards regions with lower energy, and uses a PES model to enable searching diverse stable structures and injects physical guarantee. 
Technically, we show that RSS and diffusion-based generation share the same mathematical framework of iterative structure updates, where the two methods differ only in the driving force at each step. This unification allows a combination in which the generated structure distribution  
interpolates between the learned data distribution and the Boltzmann distribution defined by the PES, resulting a concentration over diverse stable structures with low energy.
A schematic view of GSS in contrast to RSS and diffusion process is illustrated in \figref{framework}.
RSS explores the PES broadly, converging to structurally diverse local minima across different geometric patterns and guaranteeing physical validity of every returned structure. The diffusion model, on the other hand, efficiently concentrates its samples in the low-energy region near the global minimum, providing a strong prior for plausible configurations.
In contrast, GSS augments the diffusion trajectory with energy-based updates that progressively steer intermediate structures, which still carry substantial randomness, toward physically valid local minima. As sampling advances, the energy guidance increasingly dominates, ensuring convergence to local minima on the PES (see Section~\ref{sec4} for theoretical justification), while the diffusion process provides a global guide toward the low-energy region.

We evaluate GSS on representative periodic systems spanning elemental to quaternary compositions, as well as on non-periodic organic molecules.
Across all tested periodic systems, GSS consistently achieves higher structural coverage than the diffusion baseline and better energy efficiency than RSS, advancing the Pareto front defined by the two methods. GSS also recovers reference stable structures with fewer sampling trials than RSS, whereas the diffusion model saturates well below full coverage regardless of budget.
Additionally, GSS extends to elemental systems across the periodic table with consistent improvements, and demonstrates controllable conformer exploration on molecular systems.
These results establish GSS as a practical and general structure search framework that combines the physical grounding of energy-based methods with the efficiency of pretrained generative models.
In particular, GSS raises the fraction of low-energy samples from below 20\% to above 60\% on challenging systems such as carbon and silicon, and reduces the budget cost to recover all reference structures by more than an order of magnitude compared with RSS.
Notably, these gains extend to the rare ternary AlPN system, which is entirely absent from the generative model's training data, demonstrating that GSS can leverage physical guidance to generalize beyond the learned distribution.

\newcommand{\hull}{\mathrm{hull}}
\newcommand{\eV}{\,\mathrm{eV}}

\section{Results}\label{sec2}

\subsection{Overview of Generative Structure Search}
GSS is motivated by the observation that RSS and diffusion generative models can be described within a unified mathematical framework. Both methods start from randomly initialized structures and progressively update them through a sequence of iterative transformations. In general, this process takes the form:
\begin{equation}
\label{eq:general_update}
    \bfM_{i+1} = \bfM_i + \bfff_i(\bfM_i),
\end{equation}
where $\bfM_i$ denotes the structure at step $i$, and $\bfff_i(\cdot)$ is the driving force that differs between methods.

As illustrated in \figref{framework}a(left), RSS performs structure optimization by directly minimizing the potential energy. The driving force is the energy gradient:
\begin{equation}
    \bfM_{i+1} = \bfM_i - h \nabla E(\bfM_i),
\end{equation}
where $h$ denotes the step size. This update drives the system toward nearby local minimum. While RSS reliably converges to locally stable structures, exploration of different energy basins relies entirely on random initialization. Consequently, many trajectories converge to high-energy local minima, making the search for low-energy stable structures inefficient.

Diffusion models follow a different strategy. Rather than directly minimizing energy, they learn a generative process that captures the statistical distribution of 
demonstrated structures in the training data. They construct a diffusion process starting from a structure $\bfM_N$ from the dataset, which is gradually deformed and corrupted by Gaussian noise,
    $\bfM_{i-1} = \bfM_i + h \bfbb_i(\bfM_i) + g_i \sqrt{h} \bfveps,\ \bfveps \sim \clN(\bfzro,\bfI)$,
where $\bfbb_i(\cdot)$ is a vector field inducing a deterministic drift, and the scalar $g_i$ controls the noise magnitude. These two components are properly designed to make the process converge to a simple distribution that is easy to draw samples. 
Starting from such a sample as $\bfM_0$, the generation process can be performed by simulating the reverse process~\cite{anderson1982reverse,song2021score}:
\begin{equation}
    \bfM_{i+1} = \bfM_i - h \left[\bfbb_i(\bfM_i) - g_i^2 \bfss_{\theta,i}(\bfM_i)\right] + g_i \sqrt{h} \bfveps,\ \bfveps\sim\mathcal{N}(\bfzro,\bfI),
\end{equation}
where the $\theta$-parameterized score model $\bfss_{\theta,i}(\cdot)$ approximates the score function $\nabla \log p_i(\cdot)$, in which $p_i$ denotes the density function of the distribution in the $i$-th step in the forward process. 
It effectively drives each sample toward regions of higher probability density, 
analogous to the force field driving toward lower-energy regions.
As illustrated in \figref{framework}a(right), since typical training data present structures around the global energy minimum, the learned distribution concentrates on a few low-energy configurations while missing many true local minima distributed across the energy landscape (\figref{framework}c shows a concrete example on elemental carbon), 
hence not directly useful for structure search. 
Statistical generalization on new systems may also only mimic similar structures with the same geometric pattern but not physically preferred ones.

The unified form of Eq.~\eqref{eq:general_update} reveals that RSS and diffusion generation differ only in their driving forces—one grounded in physical energy, the other in a learned statistical prior. GSS integrates these two complementary mechanisms by combining them into a single update:
\begin{equation}
\label{eq:gss_sample}
  \bfM_{i+1} = \bfM_i + \alpha_i \bfff_{\mathrm{diff},i}(\bfM_i) + \beta_i \bfff_{\mathrm{RSS}}(\bfM_i),
\end{equation}
where the step-dependent coefficients $\alpha_i$ and $\beta_i$ control the relative influence of the learned generative prior and the physical energy landscape during sampling.
Through recent rapid development, the PES has been well accessed by machine-learning force fields (MLFFs)~\cite{chen2022universal,yang2024mattersim,wood2025uma}, which provide energy, force, and stress predictions with near quantum-chemical accuracy across diverse chemical and structural spaces.
The coefficients are chosen such that
$\alpha_i$ starts near one and decays toward zero, while $\beta_i$ follows the opposite trend, so that the diffusion prior dominates the early sampling steps to provide global structural guidance, and the energy-based term gradually takes over in later steps to refine structures toward nearby local minima.

As shown in Section~\ref{sec4}, the two endpoints of this schedule correspond to well-defined sampling regimes: in the limit $\alpha_i=1,\beta_i=0$, the update reduces to standard diffusion sampling from the learned structural distribution; in the limit $\alpha_i=0,\beta_i=1$, it recovers Langevin dynamics on the PES, equivalent to an RSS-style relaxation that drives samples toward local minima. Since our sigmoid schedule ends in the latter regime, the final phase of GSS acts as a relaxation step that anchors each trajectory to a local minimum of the PES, ensuring physical validity. Meanwhile, the stochastic diffusion steps during the early and intermediate phases disperse trajectories across diverse low-energy basins, so that the ensemble of final local minima covers a broader set of stable structures than unbiased relaxation alone. This coverage advantage admits a formal characterization. In Section~\ref{sec4} we show that RSS and GSS both enjoy a \emph{budget-driven coverage guarantee}, in the sense that a sufficient sampling budget recovers any stable structure with probability tending to one. The coverage of vanilla diffusion, by contrast, is intrinsically tied to its training distribution and provides no analogous budget-driven guarantee for basins absent from it.
In the context of diffusion models, such a combination is referred to as a guided generation process. There are multiple implementation strategies, including exact guidance but requires additional training~\cite{lu2023contrastive}, and training-free guidance~\cite{ye2024tfg} under certain approximations. For efficiency and flexibility, we consider a training-free formulation, and Eq.~\eqref{eq:gss_sample} is proposed for its proper balance between the two factors. See Supplementary Information Section~1.3 for further comparison and discussion.

The hybrid update of GSS leads to a characteristic search behavior illustrated in \figref{framework}a(middle): $\bfff_\mathrm{RSS}$ drives samples toward nearby local minima along the generation trajectory, while $\bfff_\mathrm{diff}$ steers a larger portion of generated structures into the low-energy region.
\figref{framework}b visualizes a concrete GSS trajectory, where the atomic score and relaxation force jointly guide the structure from a random initialization toward a stable configuration. GSS addresses this gap by combining both driving forces to achieve broader coverage of local minima with fewer wasted trials.

\subsection{Structure Search Results on Representative Periodic Systems}

\begin{figure}
\centering
\includegraphics[width=0.9\linewidth]{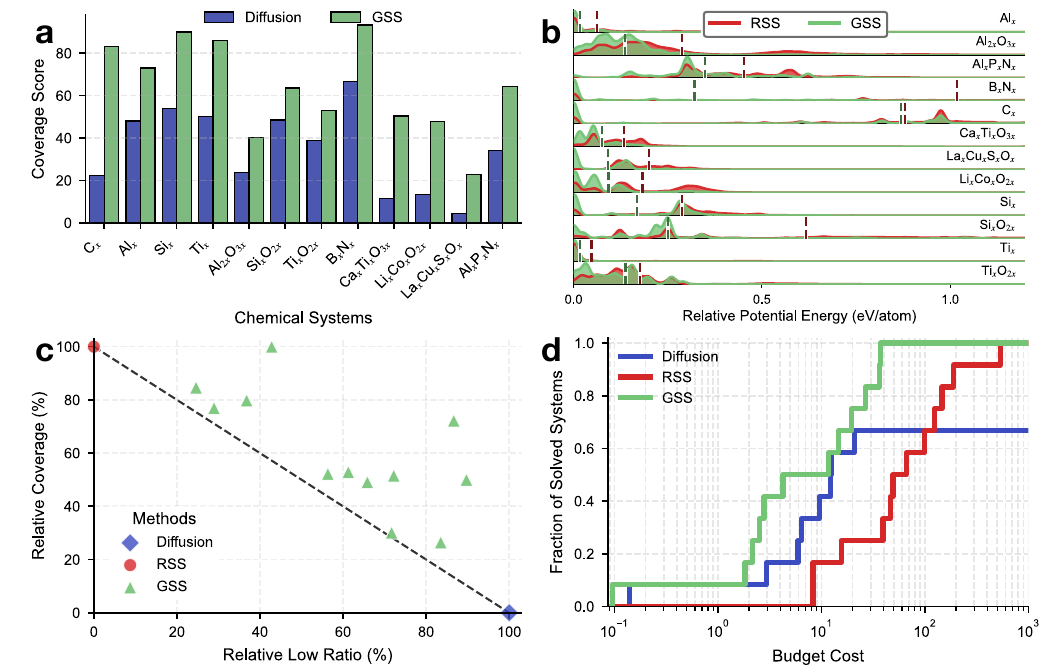}
\caption{Results of different sampling methods on twelve representative periodic systems.
\textbf{a},
The comparison of sampling coverage for each system.
The guided sampling approach achieves substantially higher coverages than the vanilla diffusion baseline when evaluated under the same number of samples. \textbf{b}, The distribution of potential energies for the sampled structures. We measure the average energy, where lower values indicate higher physical plausibility.
\textbf{c}, Illustration of the trade-off between coverage and energy efficiency. RSS and diffusion results are normalized to the top-left and bottom-right for an illustration over different systems.
All GSS results lie to the top-right of the diagonal line, indicating an advancement of Pareto front.
\textbf{d}, Budget cost comparison across methods. The $x$-axis shows the number of sampling trials normalized over the number of reference stable structures of each system; the $y$-axis shows the fraction of systems solved in the sense that all the reference stable structures have been found by a method.
Faster-rising curves indicate more efficient recovery of reference structures.
}
\label{fig:periodic_results}
\end{figure}

For a comprehensive evaluation for the structure search methods, we select twelve representative material systems ranging from elemental to quaternary compositions, including common systems with multiple known polymorphs such as $\rmC$, $\mathrm{Si}$, $\mathrm{SiO_2}$, $\mathrm{CaTiO_3}$, as well as the previously less-explored ternary $\mathrm{AlPN}$ to test the generalization performance in uncharacterized chemical spaces.
For each system, we perform an exhaustive RSS and take the generated structures as the reference set of all possible stable structures.
For a fair comparison, each method generates 1,024 structures per system. Details of the sampling procedure and post-relaxation process are provided in Supplementary Information Section~2.1.
To quantitatively evaluate the methods as structure search methods,
we introduce two metrics from two complementary perspectives (detailed in Supplementary Information Section~2.1).
\textbf{Coverage} measures the exploratory capability of a method to find diverse stable structures (similar to the recall metric).
We try to match each generated structure to each found stable structure from the reference set, and record the ratio of matched reference structures over the reference set.
\textbf{Efficiency} evaluates how effectively and rapidly the model samples energetically favorable structures (similar to the precision metric). 
We compute the average potential energy per atom, and measure the ratio of sampled structures falling within a predefined energy threshold within which local-minima structures are regarded stable (typically $0.1\eV$ above $E_\hull$).

As shown in Fig.~\ref{fig:periodic_results}a, GSS consistently achieves higher coverage of reference structures than the vanilla diffusion model across all systems.
This improvement arises because energy gradients during the diffusion trajectory steer intermediate structures away from the global minimum basin and into other metastable basins that diffusion alone cannot reach.
Meanwhile, Fig.~\ref{fig:periodic_results}b shows that the energy distribution of GSS samples concentrates more on lower energies compared with RSS, indicating improved search efficiency without sacrificing structural diversity.

These two metrics reveal a clear trade-off among methods: RSS achieves high coverage but low efficiency, whereas the diffusion model favors efficiency at the expense of limited coverage. GSS mitigates this trade-off by combining the strengths of both approaches.
To further quantify the value of the trade-off, Fig.~\ref{fig:periodic_results}c shows the \emph{relative} improvement of GSS over the two baselines. Here, we focus on the qualitative Pareto advantage, so we align the results over different systems by shifting and scaling the metrics so that RSS/diffusion has 100\%/0\% relative coverage and 0\%/100\% relative low-energy ratio.
We can observe that GSS consistently lies above the Pareto front defined by diffusion and RSS over all these systems, including the rare $\mathrm{AlPN}$ system that is scarce in the training data, indicating gains beyond what can be achieved by simply combining their sampling results.

To translate the advantages into practical benefits,
we compare the computational cost required by these methods to solve these systems. Here, we call a system is solved if a method finds all its reference stable structures.
To eliminate the influence from different difficulties in solving different systems, the cost on each system is measured by the budget cost defined as the number of sampling attempts normalized by the size of the reference stable-structure set of that system, so we can align the results over different systems (Supplementary Information Section~2.1).
Fig.~\ref{fig:periodic_results}d shows the fraction of solved systems as a function of budget cost.
It shows that GSS solves more systems than RSS and the vanilla diffusion model at all budget cost levels.
Notably, when all the systems are solved, GSS saves more than 10 times fewer budget cost than RSS.
The vanilla diffusion model can solve a few systems owing to the randomness in generating samples, but eventually only solves 8 out of the 12 systems.

\subsection{Extension over the Periodic Table}

\begin{figure}
\centering
\includegraphics[width=\linewidth]{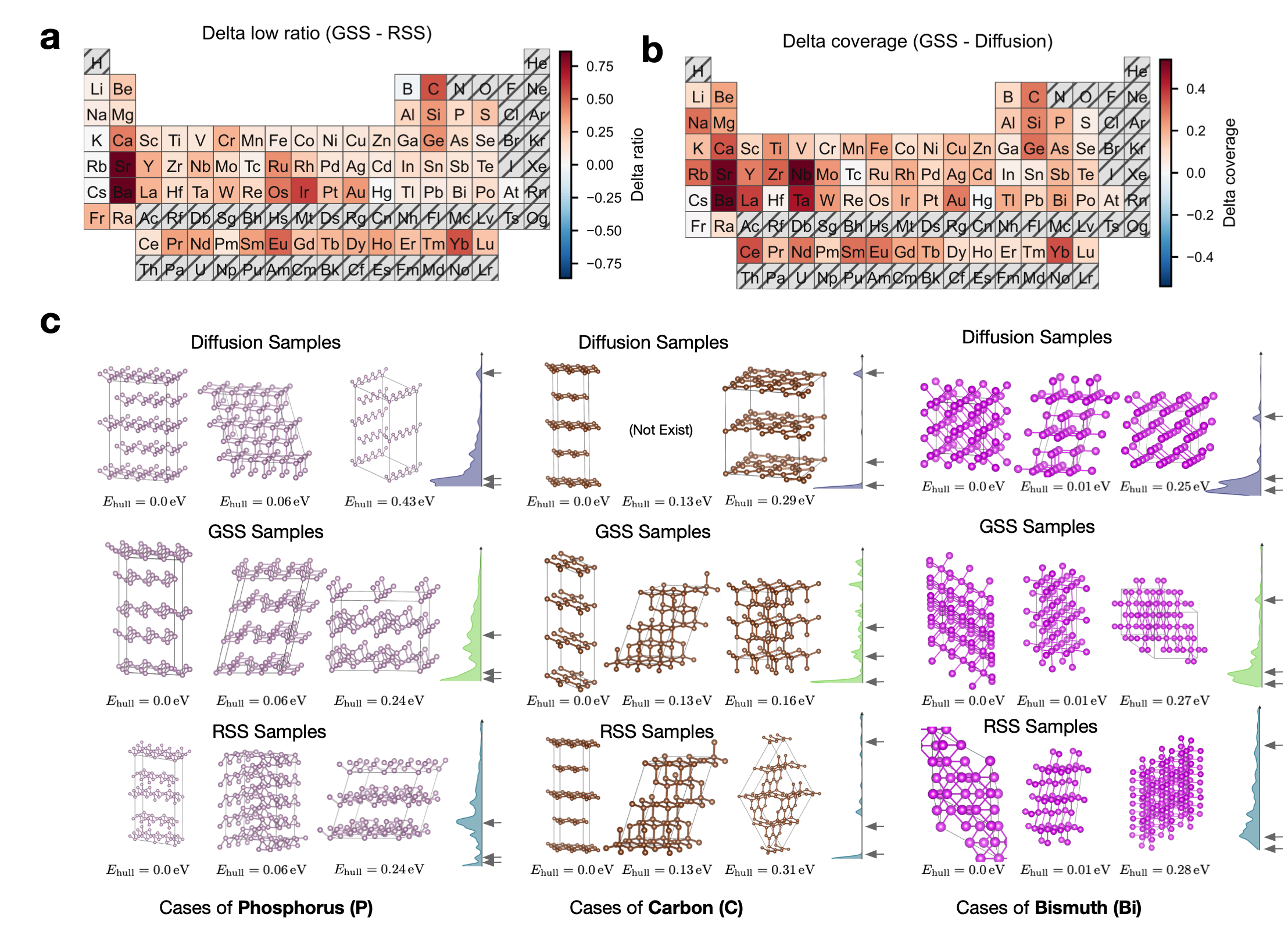}

\caption{Structure search capability demonstration on elemental crystal structures across the periodic table. \textbf{a}, Comparison of the proportion of low-energy structures produced by GSS and RSS methods. 
\textbf{b}, Comparison of the coverage over reference stable structures by structures generated by GSS and the vanilla diffusion. 
\textbf{c}, The case study on three representative elements: phosphorus ($\rmP_x$), carbon ($\rmC_x$), and bismuth ($\mathrm{Bi}_x$) ($x \in \{2,4,8\}$). 
The first two selected structures are those closest matching two well-known structures of each system, and the third is an additional high-probability structure generated by each method.
Potential energy distribution of generated structures and the potential energies of the exhibited structures are shown on the right for each case.
}
\label{fig:periodic_table}
\end{figure}

To investigate the structure search advantages of GSS across a broad range of chemical elements, we conduct a systematic experiment on elemental crystals across the periodic table.
We construct reference structure sets for elements with atomic number below 89, spanning from hydrogen to radium.
Noble gases and elements that are gaseous under normal conditions are excluded as their stable forms (monoatomic or diatomic) are trivial to identify.

As shown in \figref{periodic_table}a,b, GSS significantly improves both sampling efficiency and structural coverage over the existing methods across the periodic table. Although performance gains vary across element classes, GSS shows particularly strong improvements for alkaline earth metals (group II) and carbon group elements. These systems often exhibit complex structural landscapes with multiple competing low-energy configurations, where the vanilla diffusion model tends to concentrate on a limited subset of structures, while RSS explores the space less efficiently.

This advantage is further demonstrated through a case study on three representative elements: phosphorus ($\rmP$), carbon ($\rmC$), and bismuth ($\mathrm{Bi}$), shown in Fig.~\ref{fig:periodic_table}c.
For each method on each system, we show two generated structures each matching closest to one of two well-known stable structures of the system, 
and one additional high-probability structure sample produced by the method.
We observe that GSS and RSS can find the two well-known structures on all the systems: two white phosphorus structures for $\rmP$, the graphite and diamond structures for $\rmC$, and two $\mathrm{R\bar{3}m}$ phases for $\mathrm{Bi}$.
Although the vanilla diffusion model also finds the two structures for $\rmP$ and $\mathrm{Bi}$ from the randomness in sampling, it fails to find the diamond structure for $\rmC$, as its generated structure closest (in terms of the geometry metric that the structure matcher works with) to the diamond structure is still a layered structure. 
This reflects its limitation in systematically searching for diverse stable structures.

As for the third structure by each method, which represents the additional proposal by the searching method, GSS discovers physically realistic structures with a bit higher energy: black phosphorus for $\rmP$, hexagonal diamond (lonsdaleite) for $\rmC$, and the $\mathrm{Im\bar{3}m}$ phase for $\mathrm{Bi}$. These structures have different space groups from the two corresponding low-energy structures, demonstrating the capability to reveal qualitatively diverse stable structures.
RSS finds the same black phosphorus and $\mathrm{Im\bar{3}m}$ phases for $\rmP$ and $\mathrm{Bi}$, while for $\rmC$ it proposes hP16-carbon, a covalently bonded graphyne polymer (space group $\mathrm{P6_3/mmc}$)~\cite{hu2014covalent}. While physically plausible, this structure has a higher energy, reflecting the broad but less targeted exploration of RSS.
The vanilla diffusion-model proposals mostly have the same topology (\eg, space group) as the low-energy structures but with different parameters: a stretched layered structure for $\rmP$, a skewed and stretched graphite-like structure for $\rmC$, and a compressed structure with the same $\mathrm{R\bar{3}m}$ space group for $\mathrm{Bi}$. This reveals the limitation of statistical generalization without physical information.

Overall, GSS achieves the structure search more efficiently than RSS as its energy distribution concentrates more on the stable-structure energy regions.
In contrast to the vanilla diffusion model, the help of a PES guidance enables GSS to find diverse physically relevant structures across the periodic table. 

\subsection{Utility in Molecular Systems}

\begin{figure}[t]
\centering

\includegraphics[width=\linewidth]{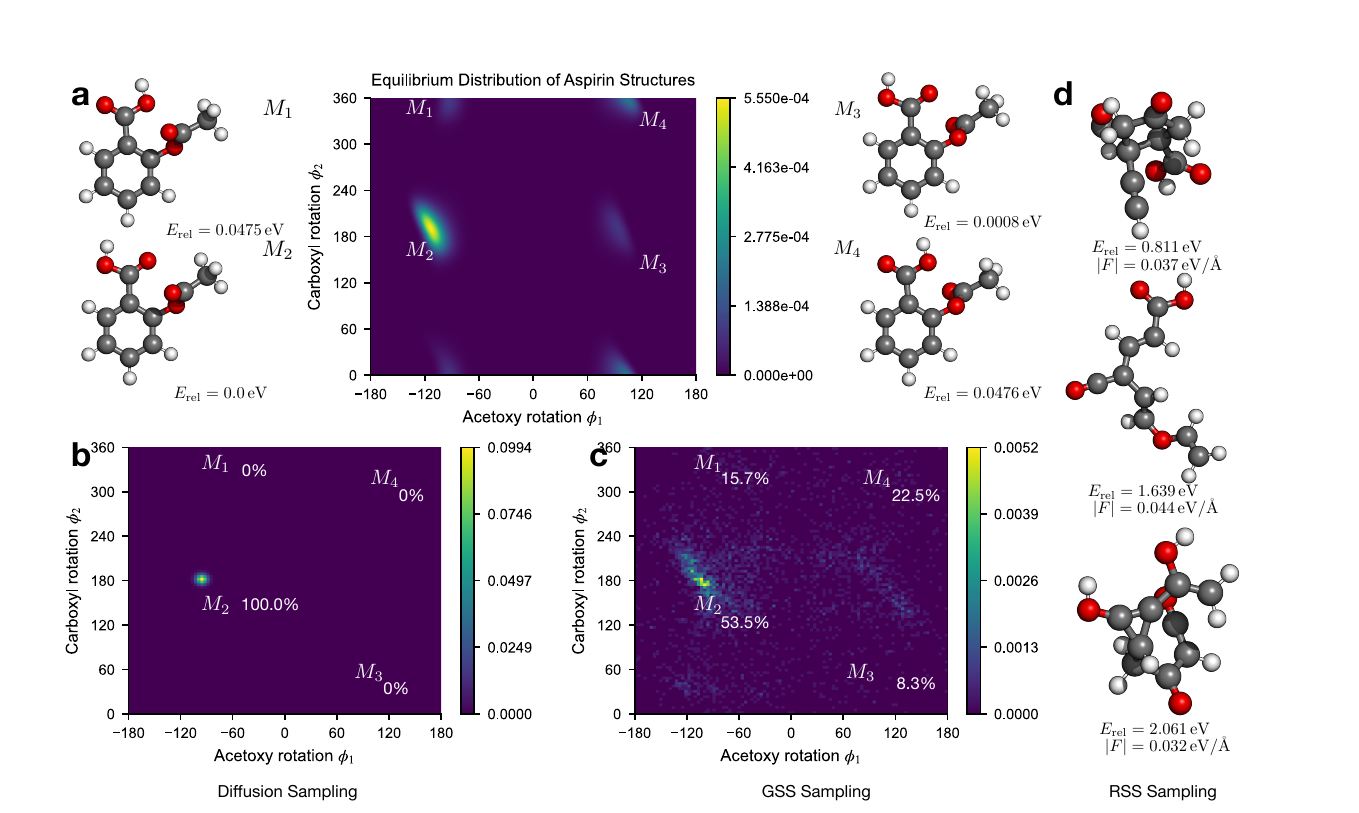}

\caption{
Structure search results for the aspirin molecule.
\textbf{a}, Normalized Boltzmann density at $T=400\,\rmK$ on its two principal degrees of freedom, \ie, the acetoxyl-benzene and carboxyl-benzene dihedral angles. 
The four stable structures $M_1$-$M_4$ are marked and visualized, together with their relative energies $E_\mathrm{rel}$ with respect to the lowest-energy structure $M_2$.
\textbf{b}-\textbf{c}, Distributions of structures produced by the vanilla diffusion model and by the GSS method, with the proportion around each stable structure marked. 
\textbf{d}, Three representative lowest-energy structures produced by RSS for the $\rmC_9\rmH_8\rmO_4$ system, with $E_\mathrm{rel}$ as well as the force scale marked.
}
\label{fig:asprin_heat}
\end{figure}

The proposed GSS method can also be extended to non-periodic molecular systems.
To examine its broader applicability, we apply GSS to a representative molecular example, the acetylsalicylic acid (aspirin, $\mathrm{C}_6\mathrm{H}_4(\mathrm{COOH})(\mathrm{O(CO)CH}_3)$).
It provides a simple yet informative test case, as its structural variability is largely dominated by two torsional angles: $\phi_{\mathrm{Ac}}$ between the acetoxyl ($\mathrm{Ac}$) and the benzene ring, and $\phi_{\mathrm{COOH}}$ between the carboxyl (-$\mathrm{COOH}$) group and the benzene ring.
In Fig.~\ref{fig:asprin_heat}a, we plot the (normalized) Boltzmann density $p(\phi_{\mathrm{Ac}},\phi_{\mathrm{COOH}})\propto\exp\left(-\frac{\mathcal{F}(\phi_{\mathrm{Ac}},\phi_{\mathrm{COOH}})}{k_\rmB T}\right)$ of aspirin structures on these two principal directions  (see Supplementary Information Section~1.2 for details).
As other degrees of freedom, \ie, bond lengths and angles, fluctuate weakly around equilibrium and are approximately independent of $\phi_{\mathrm{Ac}}$ and $\phi_{\mathrm{COOH}}$ at the relatively low temperature ($T=400\,\rmK$), the free energy $\mathcal{F}(\phi_{\mathrm{Ac}},\phi_{\mathrm{COOH}})$ is approximated by taking the fixed bond-length and angle values to construct a complete structure, which is then evaluated by a pretrained universal machine-learning force field.
The figure reveals four stable structures of aspirin, where the $M_2$ structure achieves the lowest energy.

We train a diffusion model on structures corresponding to the $M_2$ mode. As expected, Fig.~\ref{fig:asprin_heat}b shows that the 1,024 generated structures by the diffusion model all concentrate around the $M_2$ structure, with no coverage on the other three stable structures.
In contrast, with the PES guidance, GSS successfully recovers all the four stable structures, as shown in Fig.~\ref{fig:asprin_heat}c.
As for RSS, in this case, it does not generate any aspirin structure even with 2,048 trials, since it does not have a driving force towards this specific isomer; \ie, the atoms does not have to be connected into a benzene ring, an acetoxyl group and a carboxyl group. Instead, the three lowest-energy structures generated by RSS, shown in Fig.~\ref{fig:asprin_heat}d, are far from an aspirin structure. Although they are approximately energy local minima as indicated by the small force scale, their energies are well above any aspirin conformer.
In the case of non-periodic organic molecules, the PES could be more complicated with much more local minima, making RSS even harder to find a low-energy structure from randomly initialized structures.
With the oracle guidance from the diffusion model, GSS has the molecular graph topology information injected, which leads the generation process to the structure manifold of the target isomer.
This demonstrates the controllability and flexibility of GSS for broader applications.

\section{Discussion}\label{sec3}

In this work, we introduce generative structure search (GSS), a novel approach that improves both searching coverage and efficiency.
GSS incorporates both the potential energy landscape and a diffusion generative model, where the former injects physical information to guarantee proper exploration and convergence to stationary points, and the latter provides an oracle global guide towards low-energy regions. Together, GSS achieves an improvement over previous limitations by the coverage-efficiency trade-off.
Over both periodic crystalline and non-periodic molecular systems and across the periodic table, GSS achieves a comparable coverage of stable structures in much fewer sampling trials, and enables the controllability towards a specific isomer.

Beyond the methodological advances, GSS has the potential to accelerate materials and molecular discovery. By efficiently revealing possible stable structures for novel compositions, it addresses a key bottleneck in the computational pipeline from chemical composition to property prediction. Combined with the rapid progress in universal machine-learning force fields, GSS can be readily applied to previously uncharacterized systems, enabling systematic screening of candidate materials before costly experimental synthesis. For molecular systems, the ability to controllably explore conformer landscapes could benefit applications such as drug design and functional molecule discovery.

Several directions remain open for future work. The current sigmoid schedule for balancing diffusion and energy guidance is fixed before sampling; an adaptive strategy that adjusts $\alpha_i$ based on runtime diagnostics, such as the magnitude of the energy gradient or the diversity of the current sample batch, could further improve the coverage-efficiency trade-off. The guided sampling framework also naturally extends to downstream tasks beyond structure search. Combining GSS with nudged elastic band (NEB) methods could enable efficient exploration of minimum-energy pathways between known structures, where the diffusion model provides physically informed initial path guesses and energy guidance refines them toward saddle points. More broadly, applying GSS to larger-scale and more complex systems, such as protein conformational search where the energy landscape is high-dimensional and rugged, would benefit from coarse-grained or multi-fidelity guidance strategies to manage the computational cost of repeated energy and force evaluations during sampling.

\section{Methods}\label{sec4}

An open-source implementation of GSS is available at \url{https://github.com/Yifang-Qin/Generative-Structure-Search}.

\subsection{Diffusion Model for Crystal Structure Prediction}

Diffusion-based generative models \cite{ho2020denoising,song2021score,zheng2024predicting,abramson2024accurate,kim2025high,lewis2025scalable} learn to reverse a forward noising process that gradually corrupts a data sample $\bfM_N$ into a simple prior distribution over $N$ steps. As introduced in Section~\ref{sec2}, a neural score model $\bfss_{\theta,i}\approx\nabla_{\bfM_i}\log p_i(\bfM_i)$ is trained to approximate the score function at each noise level, and generation proceeds by iteratively denoising random samples following the reverse process.
Here $\bfM_i$ denotes the state variable at step $i$, whose concrete form depends on the system of interest. For non-periodic molecular systems, $\bfM=\bfR\in\mathbb{R}^{n\times 3}$ represents the Cartesian coordinates of $n$ atoms, and the forward transition is a standard Gaussian $q(\bfR_i|\bfR_N)=\clN(\sqrt{\alphab_i}\bfR_0,(1-\alphab_i)\bfI)$. For periodic crystal structures, $\bfM=(\bfX,\bfL)$ and the state space requires special treatment due to the periodicity of the crystal lattice, as we detail next.

A periodic crystal structure is described by the lattice vectors $\bfL\in\mathbb{R}^{3\times 3}$ that define the size and shape of the unit cell, and the fractional coordinates $\bfX\in[0,1)^{n\times 3}$ that specify the positions of $n$ atoms relative to the lattice vectors. Due to periodicity, the fractional coordinates are not Euclidean but lie on a hypertorus $\mathbb{T}^{3n}$. Diffusion CSP models \cite{jiao2023crystal,zeni2025generative} therefore treat $\bfX$ and $\bfL$ separately. In these methods, Brownian motion on the hypertorus leads to a wrapped Gaussian forward transition for $\bfX$ that drives the data distribution towards the uniform distribution on the hypertorus, while $\bfL$ is diffused in Euclidean space. The forward transitions take the form:
\begin{equation}
\begin{cases}
    q(\bfX_i|\bfX_N)=\clN_\tnW(\bfX_0,\sigma_i^2 \bfI, \bfI) \\
    q(\bfL_i|\bfL_N)=\clN(\sqrt{\alphab_i} \bfL_0, (1-\alphab_i) \sigma_i(n)^2 \bfI),
\end{cases}
\end{equation}
where $\clN_\tnW(\bfmu,\sigma^2\bfI,\bfI)=\sum_{\mathbf{k}\in\mathbb{Z}^{3n}}\clN(\mathbf{\mu}+\mathbf{k},\sigma^2\bfI)$ denotes the wrapped normal distribution on the unit hypertorus, $n$ is the number of atoms in the unit cell, and $\alphab_i,\sigma_i,\sigma_i(n)$ are predefined schedules that control the noise magnitude so that both $\bfX_i$ and $\bfL_i$ converge to their respective stationary distributions as $i\rightarrow 0$. A score network $\bfss_{\theta,i}(\bfM_i)$ is trained to approximate the scores of these forward transitions and used to iteratively generate samples during the reverse diffusion process.
For notational convenience, we denote the concatenated structure variable as $\bfM_i = (\bfX_i, \bfL_i)$ in the following sections.

\subsection{Guided Diffusion with Machine-Learning Force Field}

\textbf{Guided update of sampled structures.}
To incorporate physical knowledge from the potential energy surface into diffusion-based sampling, we adopt a guided sampling strategy inspired by prior work on guided diffusion. At each diffusion step $i$, the noisy structure $\bfM_i$ is updated using an external force-field model. Consistent with Section~\ref{sec2}, the energy-based guidance signal is defined as $\bfff_{\mathrm{RSS}}(\bfM_i) = -h\nabla_{\bfM_i}{E}(\bfM_i)$, which provides local information about the energy landscape.

While the score model $\bfss_{\theta,i}(\cdot)$ captures the global structure of the data distribution, the energy gradient supplies fine-grained, physically grounded corrections. Following classifier guidance \cite{dhariwal2021diffusion}, we combine these two sources into a guided score that replaces the vanilla diffusion score in the reverse sampling step:
\begin{equation}
    \label{eq:cf_guidance}
    \bfss_{\mathrm{guided},i}(\bfM_i)=\alpha_i\bfss_{\theta,i}(\bfM_i)-\beta_i\nabla_{\bfM_i}{E}(\bfM_i).
\end{equation}
Here, $\alpha_i$ and $\beta_i$ control the relative influence of the learned diffusion prior and the energy-based guidance. We set $\beta_i = 1 - \alpha_i$, so that the two contributions are complementary and sum to a consistent magnitude at every step. We schedule $\alpha_i$ smoothly over time, transitioning from exploration driven by the diffusion model to refinement guided by the energy surface. Specifically, we use a sigmoid schedule:
\begin{equation}
    \alpha_i=\mathrm{sigmoid}(\frac{i-t_{\mathrm{mid}}}{t_\mathrm{scale}})=\frac{1}{1+\exp(\frac{t_{\mathrm{mid}}-i}{t_\mathrm{scale}})},
\end{equation}
where $t_{\mathrm{mid}}$ and $t_{\mathrm{scale}}$ are pre-defined hyper-parameters that steer the guidance behavior. The sigmoid-like weight schedule ensures a gradual and stable shift between these two regimes.

\textbf{Interpretation of the guided score.}
The guided score in Eq.~\ref{eq:cf_guidance} can be understood by examining its two components.
The score model $\bfss_{\theta,i}$ is trained to approximate $\nabla_{\bfM_i}\log p_{\mathrm{data},i}(\bfM_i)$, where $p_{\mathrm{data},i}$ is the noised data distribution at step $i$.
The energy gradient $-\nabla_{\bfM_i}{E}(\bfM_i)$ corresponds to the score of the Boltzmann distribution $p_{\mathrm{energy}}(\bfM)=\frac{1}{Z}\exp(-{E}(\bfM))$, where $Z$ denotes the partition function.

The two extreme schedules are well-defined: when $\alpha_i=1$ and $\beta_i=0$, the method reduces to standard diffusion sampling from $p_{\mathrm{data}}$; when $\alpha_i=0$ and $\beta_i=1$, it recovers Langevin dynamics on the energy surface with stationary distribution $p_{\mathrm{energy}}$, equivalent to an RSS-style relaxation on the PES.
The sigmoid schedule is designed to end in the latter regime ($\alpha_i\to 0$, $\beta_i\to 1$), so that the final sampling steps act as a relaxation that anchors each trajectory to a local minimum of the PES and guarantees physical validity.
In the earlier, diffusion-dominated phase, the stochastic score updates disperse trajectories across multiple low-energy basins, enabling the ensemble of final local minima to cover a more diverse set of stable structures than an unbiased relaxation alone.

At intermediate steps, the energy term evaluates $E(\bfM_i)$ on the noisy intermediate rather than on the underlying clean structure $\bfM_N$, introducing an approximation standard in classifier guidance \cite{dhariwal2021diffusion} and training-free guidance \cite{ye2024tfg}.
The sigmoid schedule mitigates this gap by activating the energy guidance only in the late, low-noise stage, where the intermediate is already close to a clean structure and its energy closely reflects that of the final sample.
One can further refine this by evaluating the energy on an estimate of the clean structure $\mathbb{E}[\bfM_{N|i}]$ obtained from the score model, which we refer to as the $\bfM_N$-guided variant and discuss in Supplementary Information Section~1.3.

Under this view, the guidance schedule acts as an annealing mechanism that progressively increases the influence of the true potential energy, enabling a smooth transition from model-driven generation to physically informed refinement.

\textbf{Leveraging the energy function in periodic sampling.}
Both guidance methods rely on the energy gradients of periodic structures $\bfM=(\bfX,\bfL)$. However, major MLFFs typically output forces and stresses rather than energy gradients w.r.t fractional coordinates $\bfX$ and lattice vectors $\bfL$ in the form required by periodicity-oriented diffusion models. Integrating these force-field outputs into the diffusion sampling process therefore requires careful transformations to convert forces and stresses into consistent updates of fractional coordinates and lattice vectors.

For coordinate guidance, atomic forces are obtained as the negative gradient of the energy with respect to Cartesian positions, $\bfF_{\mathrm{cart}}(\bfM)=-\nabla_{\bfR}{E}(\bfM)$, where $\bfR=\bfL^T\bfX$ denotes the Cartesian coordinates of atoms. Using the chain rule, the corresponding gradient with respect to the fractional coordinate of atom $j$ can be derived as:
\begin{equation}
    \nabla_{\bfX_j}{E}(\bfM) = \nabla_{\bfX_j}\bfR_j\,\nabla_{\bfR_j}{E}(\bfM) = -\bfL\bfF_{\mathrm{cart},j},
\end{equation}
where $j$ indexes atoms in the unit cell.
For lattice guidance, the total stress tensor is defined as:
\begin{equation}
    \bfsigma_{\mathrm{total}} = \frac{1}{|\det(\bfL)|}\,\nabla_{\bfeta}{E}(\bfX,\bfL(\bfI+\bfeta))\bigg|_{\bfX},
\end{equation}
where $|\det(\bfL)|$ represents the unit cell volume.
Based on this definition, the gradient of the energy with respect to the lattice parameters follows as
\begin{equation}
    \nabla_{\bfL}{E}(\bfM)=-|\det(\bfL)|\bfL^{-T}\bfsigma_{\mathrm{total}}+\bfX^T\bfF_{\mathrm{cart}}.
\end{equation}
In practice, most MLFF toolkits such as ASE \cite{larsen2017atomic} output the virial stress rather than the total stress. Since our lattice guidance requires $\nabla_{\bfL}{E}(\bfM)$, we need to convert between the two conventions. The virial stress is defined as:
\begin{equation}
    \bfsigma_{\mathrm{virial}} = -\frac{1}{|\det(\bfL)|}\,\nabla_{\bfeta}{E}(\bfL^{-T}\bfR,\bfL(\bfI+\bfeta))\bigg|_{\bfR} = \bfsigma_{\mathrm{total}}-\frac{1}{|\det(\bfL)|}\bfF_{\mathrm{cart}}\bfR.
\end{equation}
Substituting this relation, the lattice gradient expressed in terms of the virial stress simplifies to:
\begin{equation}
    \nabla_{\bfL}{E}(\bfM)=-|\det(\bfL)|\bfL^{-T}\bfsigma_{\mathrm{virial}}.
\end{equation}

These gradients with respect to fractional coordinates $\bfX$ and lattice parameters $\bfL$ provide a consistent way to inject energy information from force-field models into the guided diffusion sampling process.

\subsection{Coverage Guarantees from Stochastic Guidance}
The two regimes identified in the previous subsection, namely diffusion-dominated exploration and energy-dominated relaxation, can be tightened into formal coverage statements that distinguish GSS from both RSS and vanilla diffusion.
We focus on the periodic case $\bfM=(\bfX,\bfL)\in\mathcal{M}:=\mathbb{T}^{3n}\times\mathbb{R}^{3\times 3}$; the molecular case is entirely analogous.
Let $\{\bfM^{(k)}\}_{k=1}^K$ denote the local minima of the PES $E$. For each $\bfM^{(k)}$, we define its basin of attraction in terms of the limiting behavior of deterministic energy relaxation,
\begin{equation}\label{eq:basin}
    \mathcal{B}_k \;=\; \big\{\bfM\in\mathcal{M}\,:\,\mathrm{Relax}_E(\bfM)=\bfM^{(k)}\big\},
\end{equation}
where $\mathrm{Relax}_E$ denotes the limiting point of the $-\nabla E$ gradient flow, or any equivalent deterministic local relaxation procedure. We say that a sampler returns $\bfM^{(k)}$ whenever its output, after such relaxation, falls into $\mathcal{B}_k$.

A sampler is said to provide a \emph{budget-driven coverage guarantee} for $\bfM^{(k)}$ if a single independent trial enters $\mathcal{B}_k$ with probability at least some $\rho_k>0$. By independence, $n$ independent trials then recover $\bfM^{(k)}$ with probability at least $1-(1-\rho_k)^n$, which tends to one as $n\to\infty$. This notion captures the coverage mechanism of structure search: any nonzero single-trial probability of entering the target basin is amplified to unity by a sufficiently large sampling budget.

We write $p_\mathrm{init}$ for the distribution of $\bfM_0$ used by RSS, and $p_{\mathrm{gss},i}$ for the marginal of $\bfM_i$ produced by the guided update in Eq.~\ref{eq:cf_guidance}, analogous to $p_{\mathrm{data},i}$ for the vanilla diffusion sampler. In practice, $p_\mathrm{init}$ is realized by AIRSS-style random initialization \cite{pickard2006high,pickard2011ab}, which applies physically motivated constraints such as a minimum interatomic distance and a bounded initial density range to improve relaxation efficiency and search stability. These constraints carve out unphysical regions of $\mathcal{M}$ but preserve broad coverage of physically relevant basins, so that $p_\mathrm{init}(\mathcal{B}_k)>0$ for every $\bfM^{(k)}$ of interest.

\begin{proposition}[RSS coverage]\label{prop:rss}
For any stable structure $\bfM^{(k)}$ with $p_\mathrm{init}(\mathcal{B}_k)>0$, after $n$ independent RSS trials,
\begin{equation}
    \Pr\!\big(\exists\,\text{trial entering }\mathcal{B}_k\big) \;\geq\; 1-\big(1-p_\mathrm{init}(\mathcal{B}_k)\big)^n \;\xrightarrow{n\to\infty}\;1.
\end{equation}
That is, RSS provides a budget-driven coverage guarantee for $\bfM^{(k)}$ with rate $\rho_k=p_\mathrm{init}(\mathcal{B}_k)$.
\end{proposition}

The argument is immediate: each trial draws $\bfM_0\sim p_\mathrm{init}$ and applies deterministic relaxation, which by definition of the basin converges to $\bfM^{(k)}$ whenever $\bfM_0\in\mathcal{B}_k$. Independence across trials then yields the stated bound.

Diffusion sampling does not enjoy a similar guarantee. Its training data are typically concentrated near the global energy minimum on each system, so stable structures $\bfM^{(k)}$ whose energies depart appreciably from the global minimum receive vanishing training mass in their basins. To make this precise we consider the exact-sampler population limit, in which the score model is exact and the diffusion sampler returns the data distribution: $\bfM_N\sim p_\mathrm{data}$. Under this idealization the following holds.

\begin{proposition}[Exact diffusion lacks a budget-driven coverage guarantee outside training]\label{prop:diff}
Let $q_\mathrm{diff}=p_\mathrm{data}$ denote the terminal distribution of the exact diffusion sampler. For any basin $\mathcal{B}_k$ with $p_\mathrm{data}(\mathcal{B}_k)=0$,
\begin{equation}
    \Pr\!\big(\exists\,\text{diffusion trial entering }\mathcal{B}_k\big)=0
\end{equation}
for every sampling budget $n$. Consequently, exact diffusion does not provide a budget-driven coverage guarantee for any basin absent from the training distribution.
\end{proposition}

\begin{remark}\label{rmk:leakage}
Proposition~\ref{prop:diff} should not be read as claiming that a finite, trained diffusion model can never output structures outside $\mathrm{supp}(p_\mathrm{data})$. Finite-capacity score networks, discretization error in the reverse process, residual terminal noise, and out-of-distribution extrapolation may all assign nonzero probability to regions with $p_\mathrm{data}(\mathcal{B}_k)=0$. Such leakage, however, is not a controlled basin-search mechanism: the diffusion objective itself provides no lower bound on $q_\mathrm{diff}(\mathcal{B}_k)$ for basins absent from training, so the success probability cannot be reliably amplified by increasing $n$. This is consistent with the empirical observation in Section~\ref{sec2} that the vanilla diffusion model saturates well below full coverage on metastable structures.
\end{remark}

\begin{proposition}[GSS coverage]\label{prop:gss}
For any stable structure $\bfM^{(k)}$, suppose there exists an intermediate step $i^*\in(0,N)$ such that
\begin{enumerate}
    \item[\textnormal{(a)}] $p_{\mathrm{gss},i^*}(\mathcal{B}_k)>0$, and
    \item[\textnormal{(b)}] for all $i\geq i^*$ the schedule satisfies $\beta_i\gg\alpha_i$, so that the GSS update on $[i^*,N]$ is dominated by $\bfff_\mathrm{RSS}$ and behaves as deterministic relaxation along $-\nabla E$.
\end{enumerate}
Then after $n$ independent GSS trials,
\begin{equation}
    \Pr\!\big(\exists\,\text{trial entering }\mathcal{B}_k\big) \;\geq\; 1-\big(1-p_{\mathrm{gss},i^*}(\mathcal{B}_k)\big)^n \;\xrightarrow{n\to\infty}\;1.
\end{equation}
Hence GSS provides a budget-driven coverage guarantee for $\bfM^{(k)}$ with rate $\rho_k=p_{\mathrm{gss},i^*}(\mathcal{B}_k)$.
\end{proposition}

The argument is again immediate: by~(a), a single trajectory has probability $p_{\mathrm{gss},i^*}(\mathcal{B}_k)$ of being in $\mathcal{B}_k$ at step $i^*$; by~(b), the subsequent updates on $[i^*,N]$ relax deterministically along $-\nabla E$ and therefore converge to $\bfM^{(k)}$ once inside the basin. Independence across trials gives the bound.

Condition~(b) is precisely the regime entered by our sigmoid schedule once $i$ crosses $t_\mathrm{mid}$, so its existence is built into the GSS design. Condition~(a) is the substantive requirement: it asserts that the early, stochastic phase of GSS spreads trajectories broadly enough to enter $\mathcal{B}_k$ with nonzero probability. The intermediate marginals of vanilla diffusion are likewise smoothed by Gaussian noise, so this condition is in fact also satisfied by the intermediate states of vanilla diffusion. The decisive difference therefore lies in the second phase: vanilla diffusion completes its trajectory by data-score denoising, which pulls samples back toward $\mathrm{supp}(p_\mathrm{data})$ regardless of which basin the intermediate state had visited, whereas GSS replaces this denoising with energy-dominated relaxation, which locks any visited basin into its local minimum.

Propositions~\ref{prop:rss}--\ref{prop:gss} together formalize the qualitative picture in Section~\ref{sec2}. The three samplers can be summarized as follows: RSS combines broad initialization with energy relaxation; vanilla diffusion performs training-distribution sampling without any basin-level guarantee; GSS combines stochastic exploration with energy relaxation. In both RSS and GSS, the energy-relaxation stage is what converts a single-trial probability of entering a basin into convergence to the corresponding local minimum, yielding a budget-driven coverage guarantee for every physically accessible basin. The coverage of vanilla diffusion, by contrast, is intrinsically tied to its training distribution.

\subsection{Experimental Settings}

\subsubsection{Pretrained CSP and Force Field Models}
We follow the score-network design used in MatterGen \cite{zeni2025generative}, which models the diffusion of fractional atomic coordinates $\bfX_t$ and a six d.o.f. lattice parameter $\bfL_t$. The model is randomly initialized and trained on the CSP task using the Alex-MP-20 dataset \cite{zeni2025generative}, a combined collection of 545,048 DFT-relaxed structures from the Materials Project \cite{jain2013commentary} and Alexandria \cite{schmidt2022large}. The obtained MatterGen checkpoint produces score predictions conditioned on a chemical formula, perturbed atomic positions, cell parameters, and the diffusion timestep.

For the machine learning force field (MLFF) model, we adopt the MatterSim \cite{yang2024mattersim} model to obtain the guidance information during diffusion generation. Specifically, all sampling and reference calculations involves energy information rely on the publicly released MatterSim-v1.0.0-5M checkpoint \cite{matsim_ckpt}.

\subsubsection{Evaluation Datasets}
\textbf{Representative Crystalline Systems.}
We evaluate our approach and a diffusion baseline across a diverse set of periodic systems, including elemental ($\mathrm{Al}_x$, $\mathrm{C}_x$, $\mathrm{Si}_x$, $\mathrm{Ti}_x$), binary ($\mathrm{Al}_{2x}\mathrm{O}_{3x}$, $\mathrm{Si}_{x}\mathrm{O}_{2x}$, $\mathrm{B}_{x}\mathrm{N}_{x}$, $\mathrm{Ti}_{x}\mathrm{O}_{2x}$), ternary ($\mathrm{Ca}_{x}\mathrm{Ti}_{x}\mathrm{O}_{3x}$, $\mathrm{Li}_{x}\mathrm{Co}_{x}\mathrm{O}_{2x}$, $\mathrm{Al}_{x}\mathrm{P}_{x}\mathrm{N}_{x}$), and quaternary ($\mathrm{La}_{x}\mathrm{Cu}_{x}\mathrm{S}_{x}\mathrm{O}_{x}$) compositions. We have also included two randomly chosen systems ($\mathrm{Al}_{x}\mathrm{P}_{x}\mathrm{N}_{x}$ and $\mathrm{La}_{x}\mathrm{Cu}_{x}\mathrm{S}_{x}\mathrm{O}_{x}$) to further assess model generalization.

For each composition, reference structures are generated using RSS that start from series of initial proposal structures generated by AIRSS package \cite{pickard2006high,pickard2011ab}.
Geometry relaxations of the RSS use the FIRE optimizer \cite{bitzek2006structural} and the convergence criterion is set as $f_{\max}\le0.01 \,\mathrm{eV/\AA}$. For all the initial proposals, only structures converged within 1,000 relaxation steps are retained. We then remove duplicates and filter out high-energy structures with $E_\mathrm{hull} > 1.0\,\mathrm{eV}$ to ensure stability and physical relevance. This extensive RSS procedure aims to exhaustively sample the energy landscape broadly and capture a wide set of low-energy configurations.

\textbf{Elemental Periodic Systems}
To further test the method’s ability to explore structural space, we generate reference RSS datasets for all elemental solids with atomic number below 89. It is noteworthy that gas-phase atoms like $\mathrm{H}$, $\mathrm{N}$, and $\mathrm{O}$, as well as noble-gas atoms like $\mathrm{He}$ and $\mathrm{Ne}$ are excluded from the searching process. Processing of the elemental systems follows the same filtering and de-duplication steps as above.


\bibliography{sn-bibliography}


\end{document}